\documentclass{article}

\usepackage{PRIMEarxiv}
\usepackage{verbatim}
\usepackage{mhchem}
\usepackage{float}
\usepackage[utf8]{inputenc} 
\usepackage[T1]{fontenc}    
\usepackage{hyperref}       
\usepackage{url}            
\usepackage{booktabs}       
\usepackage{amsfonts}       
\usepackage{nicefrac}       
\usepackage{microtype}      
\usepackage{lipsum}
\usepackage{fancyhdr}       
\usepackage{graphicx}       
\usepackage{listings}
\usepackage{xcolor} 
\usepackage{amsmath}
\usepackage{amssymb}
\usepackage[russian, english]{babel}
\usepackage{enumitem} 

\newcommand{\note}[1]{\textit{#1}}

\graphicspath{{media/}}     

\title{
PereStruct: Multimodal Semantic Assembly for Robust Historical Document Parsing
}

\author{
  Maksim Shandybo$^{1,2}$, Ivan Bespalov$^{1}$, Daniil Yefimov$^{2,3}$, Marina Kosheleva$^{2}$, Alexander Loukianov$^{4}$\\
  $^{1}$IGIC RAS, Moscow, Russian Federation\\
  $^{2}$Yandex Cloud, Moscow, Russian Federation\\
  $^{3}$National University of Science and Technology MISIS, Moscow, Russian Federation\\
  $^{4}$Nekrasov Central Universal Scientific Library, Moscow, Russia
}

\begin{document}
\maketitle

\begin{abstract}
Parsing historical documents with complex, non-standard layouts remains a fundamental bottleneck in large-scale archival digitization. Unlike modern typography, historical newspapers exhibit severe physical degradation and highly irregular page structures that confound even state-of-the-art vision-language models, presenting severe out-of-distribution challenges. We address this gap with an automated pipeline specifically designed for parsing historical newspapers, documents characterized by particularly intricate multi-column layouts. Our approach combines a fine-tuned YOLO architecture for layout analysis and block detection, trained on 1,426 fully human-annotated scanned pages, with a novel semantic assembly module that reconstructs articles by jointly modeling lexical-semantic similarity via TF-IDF, visual embeddings from our fine-tuned YOLO, and geometric layout constraints. This multi-modal integration yields state-of-the-art performance, achieving an F1 score of 0.904 on block-to-article mapping. Notably, end-to-end evaluation against vision-language models (Qwen3.6-35B-A3B and Qwen3.6-Plus) demonstrates that PereStruct achieves substantially higher fidelity (BLEU $\approx$0.96 vs $\approx$0.34), validating that modular architectures excel where generic VLMs fail on complex historical layouts. To support reproducibility and advance research in this domain, we release both the training corpus of 599 annotated pages and a curated \textbf{PereStruct} benchmark of 93 pages with expert-verified ground-truth block-to-article mappings. This framework establishes a robust foundation for high-fidelity digitization and semantic reconstruction of complex archival materials.
\end{abstract}


\note{Code and data are available at: \url{https://github.com/makSShandybo/PereStruct}}

\section{Introduction}

Processing historical documents with unconventional layouts remains a key obstacle to large-scale archive digitization. Although recent vision-language and document understanding systems have made strong progress on modern structured documents, they still struggle when faced with historical pages that combine degraded scans, irregular typography, dense text, and unstable page geometry. These limitations are especially pronounced in newspaper archives, where the page is not merely a container for text but a visually dense composition of articles, headlines, captions, advertisements, tables, and illustrations arranged according to conventions that vary across time, publication, and even within a single issue.
\cite{
zhang_document_2026, 
philips_historical_2020, 
banerjee_making_2024}

Historical newspapers are particularly challenging because their layouts often depart from modern linear reading assumptions. Multi-column article flows, nested headlines, discontinuous text regions, and tightly packed blocks create ambiguous reading order and make article boundary detection difficult even for advanced models. At the same time, physical degradation, including fading, stains, skew, bleed-through, torn edges, and scanning artifacts, further amplifies the out-of-distribution gap between archival pages and the cleaner document images on which many current systems are trained. As a result, document parsing in this setting is not only an OCR problem, but a multi-stage reconstruction task that requires detecting layout units, recovering their relations, and assembling them into semantically coherent articles.
\cite{ 
quattrini_gat_2024,
fleischhacker_text_2025}

Recent literature increasingly recognizes that layout analysis is the critical first step in this process. Survey work on document parsing emphasizes that modular pipelines still dominate because end-to-end models often lack reliable layout grounding, struggle with high-density text, and fail to preserve spatial structure in complex documents. In parallel, studies on historical document processing and newspaper collections show that progress depends on robust page segmentation, annotation quality, and domain-specific adaptation to archive materials, rather than generic document benchmarks. This is particularly relevant for newspaper digitization, where the problem is not just recognizing characters but identifying which text fragments belong together, in what order they should be read, and where one article ends and another begins.
\cite{ 
kontonasios_automating_2026,
satheesan_toward_2022}

Several recent works point to the same core obstacle: historical document understanding requires models that are sensitive to both visual layout and semantic context. For example, document parsing research highlights the importance of integrating layout detection with content extraction and multi-modal signals, while historical newspaper studies stress the role of metadata enrichment and structured reconstruction for improving retrieval and searchability. Likewise, recent efforts on historical newspaper corpora demonstrate that annotated datasets and block-level supervision are essential for learning robust page-to-article mappings under noisy archival conditions. These findings suggest that semantic reconstruction must be treated as a structured inference problem rather than a purely textual one.
\cite{ 
kostelnik_textbite_2025,
ali_computer_2023}

In this work, we address this gap with an automated pipeline designed specifically for historical newspapers characterized by intricate multi-column layouts and severe visual noise. Our method combines layout analysis, block detection, and semantic assembly to recover article structure from scanned pages, using both geometric constraints and semantic similarity cues to resolve ambiguous block associations. By training on fully human-annotated pages and evaluating on expert-verified ground truth, we aim to advance high-fidelity parsing of archival newspapers and provide resources that support reproducible research in historical document understanding.

\section{PereStruct Architecture}
\label{sec:headings}

The PereStruct architecture comprises three sequential stages:

\begin{itemize} 
 \item[\textbf{(1)}] \textbf{Layout Detection.} We employ a fine-tuned YOLO model to detect text and figure blocks and extract bounding box coordinates and their types from scanned newspaper pages.

\item[\textbf{(2)}] \textbf{Text Extraction.} OCR is applied to the detected regions to obtain raw text fragments corresponding to each layout block with subsequent correction using an LLM.

\item[\textbf{(3)}] \textbf{Semantic Assembly.} A classification model reconstructs articles by jointly leveraging three input signals: TF-IDF vectors encoding lexical-semantic similarity, visual embeddings from the fine-tuned YOLO encoder, and geometric coordinates governing spatial relationships. This multi-modal integration yields sets of semantically coherent articles from disjointed text blocks.
\end{itemize}

\begin{figure}[h!]
    \centering
    \includegraphics[width=\linewidth]{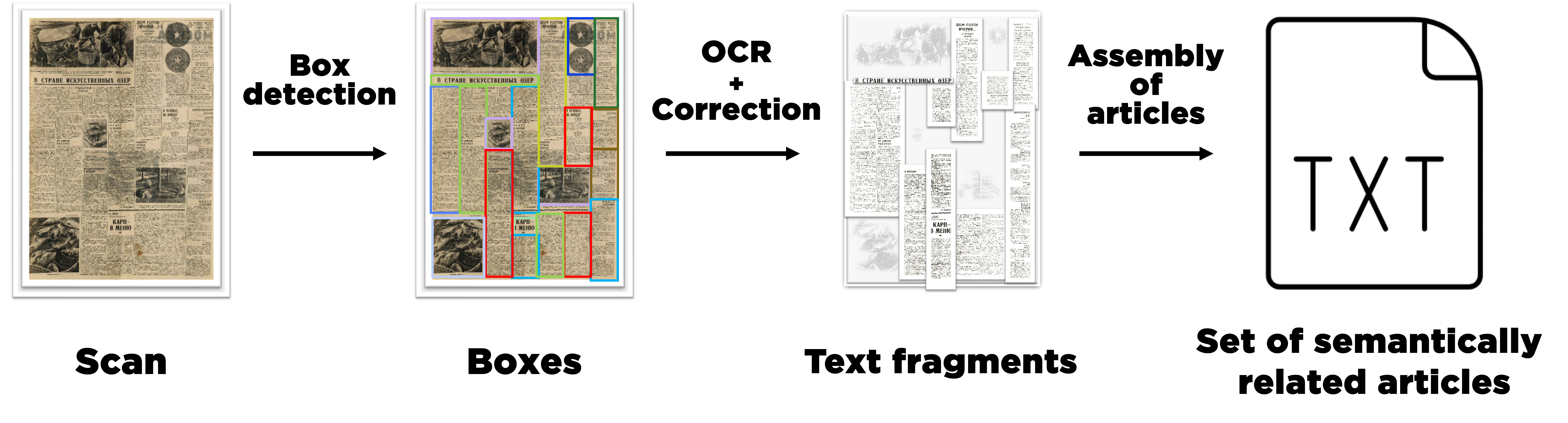} 
    \caption{\textbf{PereStruct} architecture visualization.}
\end{figure}

\section{Data}

To support reproducible research and enable further development in historical document parsing, we release two complementary open-source datasets derived from Soviet newspapers spanning 1934--1969.

\subsection{Layout Detection Corpus}
We curate a large-scale training dataset comprising 1,426 high-resolution scanned pages from the N.A. Nekrasov Library collection, annotated for layout detection in standard YOLO format. These annotations were produced in collaboration with the Yandex Crowd team, utilizing a rigorous multi-stage quality control protocol to ensure consistent bounding box placement across the heterogeneous visual styles of the era. For each page, the dataset provides normalized bounding box coordinates $(x_{\text{center}}, y_{\text{center}}, \text{width}, \text{height})$ alongside categorical labels spanning four classes: \textit{Title}, \textit{Plain Text}, \textit{Figure}, and \textit{Figure Caption}. 

While our layout detection model was trained on the full corpus of 1,426 images to capture the full spectrum of observed complexity, due to copyright restrictions, we publicly release a subset comprising 599 annotated pages. This corpus provides a robust foundation for training domain-specific detectors and evaluating performance on irregular multi-column flows and varying degradation patterns.

\subsection{PereStruct: Expert Benchmark for Article Reconstruction}
Additionally, we introduce a high-precision evaluation benchmark featuring expert-verified annotations for both layout detection and semantic article assembly. The second stage of our pipeline (article assembly) was developed and validated using 110 expert-verified pages. However, due to copyright reasons, the public release of the \textbf{PereStruct} benchmark consists of 93 high-precision annotated pages. 

Distinct from the general training corpus, this subset underwent rigorous triple-check validation by domain specialists to achieve near-flawless ground truth. Beyond bounding boxes, the benchmark includes \textit{block-to-article} mappings that explicitly encode the logical structure of newspaper pages. Each entry provides complete article reconstruction context: ground-truth articles are represented as structured JSON objects grouping disjointed text blocks into semantically coherent reading sequences. Each block is enriched by OCR-extracted text content subjected to post-correction using a large language model, enabling end-to-end evaluation of both geometric layout detection and semantic content reconstruction within a unified framework.

\begin{figure}[h!]
    \centering
    \includegraphics[width=0.50\linewidth]{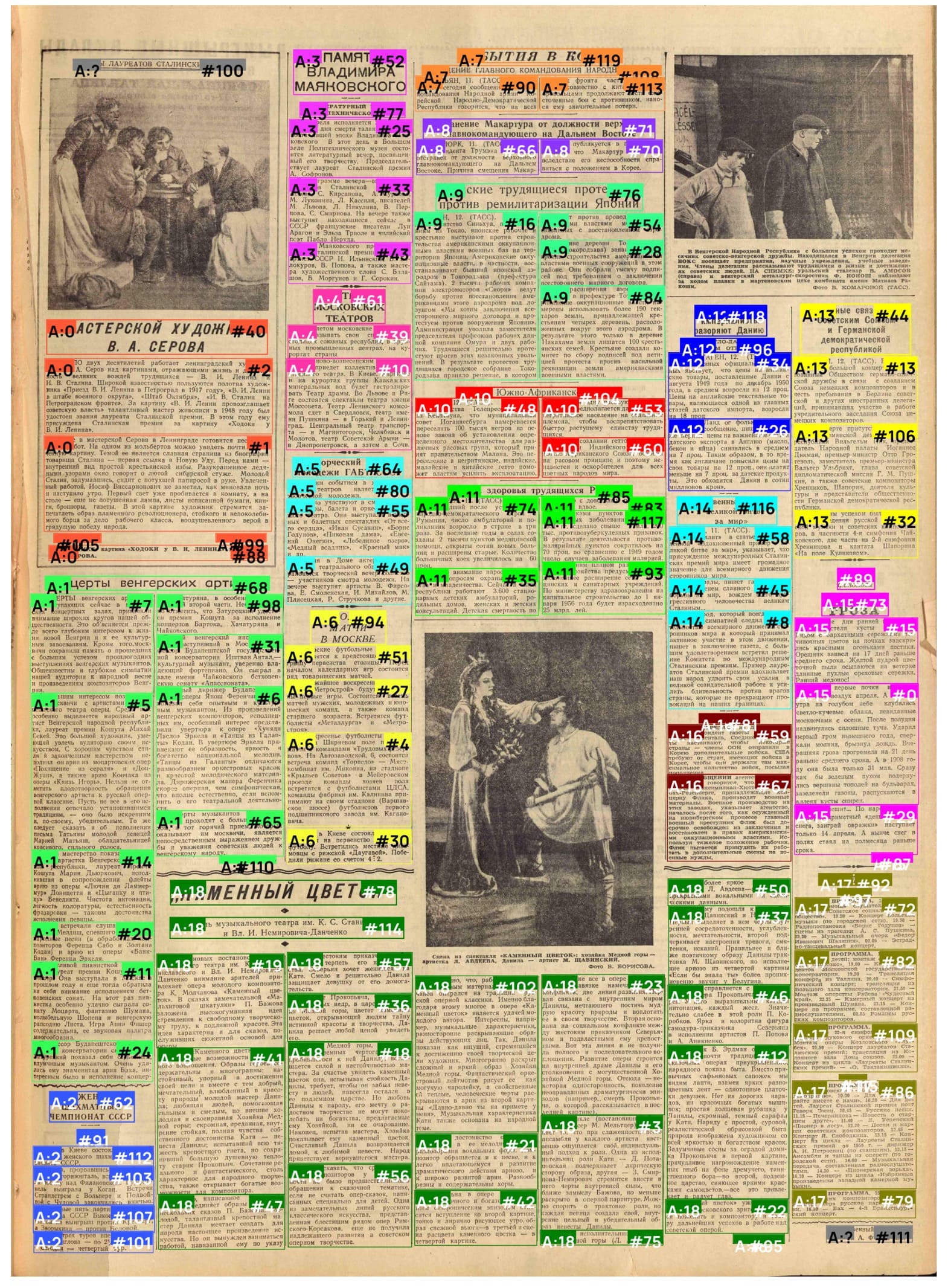} 
    \caption{\textbf{PereStruct} Benchmark block-to-article mappings visualization.}
\end{figure}

\section{Layout Detection}
\subsection{YOLOv10 Pre-training}

We initialized our layout detection pipeline using the pre-trained weights from the DocLayout-YOLO framework, specifically the checkpoint fine-tuned on DocStructBench at 1024\,px resolution \cite{doclayoutyolo2024}. While DocLayout-YOLO delivers robust general-purpose document understanding across contemporary journals and newspapers, we observed a pronounced performance degradation when directly applying it to our corpus of historical newspapers. This decline stems from a severe domain shift: Soviet-era publications feature disproportionately large page dimensions, highly non-standardized and heterogeneous layouts, irregular column configurations, and a wide variety of period-specific typographic styles and image placements that fall far outside the distribution of modern document datasets.

To bridge this domain gap, we initialize the YOLOv10 backbone from the DocLayout-YOLO checkpoint (DocStructBench, 1024 px) and adapt it using the pre-training corpus described in Section 3.1. We preserve the original architecture but restrict the classification head to the four task-critical categories defined in our taxonomy: \textit{Title}, \textit{Plain Text}, \textit{Figure}, and \textit{Figure Caption}. This focused output space aligns directly with downstream article extraction objectives while the stratified temporal coverage (1934--1969) ensures robust feature learning across the era's heterogeneous visual styles.

The impact of domain-specific pre-training is quantitatively summarized in Table~\ref{tab:yolo_comparison}, evaluated on a held-out test set of 72 pages. Compared to the original DocLayout-YOLO baseline, our pre-trained YOLOv10 model yields substantial gains across all metrics. The overall mAP50 increases from 0.698 to 0.845, while mAP50--95 improves from 0.605 to 0.750. Class-level improvements are particularly pronounced for structurally ambiguous elements: \textit{Title} detection sees a +0.230 jump in mAP50, and \textit{Figure Caption} recognition improves by +0.166 (mAP50) and +0.123 (mAP50--95). These results confirm that exposure to historical typography and layout conventions successfully recalibrates the model's feature representations for our target domain.
\begin{table}[h]
    \centering
    \caption{Comparative performance metrics of the source DocLayout-YOLO and pre-trained YOLOv10 models on the pre-train dataset test set.}
    \label{tab:yolo_comparison}
    \begin{tabular}{lcccccc}
        \toprule
        \textbf{Class} & \textbf{Sample Count} & \multicolumn{2}{c}{\textbf{DocLayout-YOLO}} & \multicolumn{2}{c}{\textbf{Pre-trained YOLOv10}} \\
        \cmidrule(lr){3-4} \cmidrule(lr){5-6}
                       &                       & \textbf{mAP50} & \textbf{mAP50-95} & \textbf{mAP50} & \textbf{mAP50-95} \\
        \midrule
        All            & 8202                  & 0.698          & 0.605             & \textbf{0.845} & \textbf{0.750}   \\
        Title          & 1195                  & 0.634          & 0.528             & \textbf{0.864} & \textbf{0.734}   \\
        Plain Text     & 6714                  & 0.795          & 0.689             & \textbf{0.900} & \textbf{0.838}   \\
        Figure         & 155                   & 0.837          & 0.810             & \textbf{0.923} & \textbf{0.914}   \\
        Figure Caption & 138                   & 0.525          & 0.391             & \textbf{0.691} & \textbf{0.514}   \\
        \bottomrule
    \end{tabular}
\end{table}

\subsection{YOLOv10 Fine-tuning}

Following domain adaptation, we perform targeted fine-tuning on the high-precision benchmark detailed in Section 3.2. We partition the 110 expert-verified pages into 100 for fine-tuning and 10 pages held out strictly for final evaluation. Given the near-flawless annotation quality achieved through triple-check validation, we utilize this subset to refine decision boundaries and suppress residual false positives on the most challenging layout configurations.

Table~\ref{tab:yolo_benchmark_comparison} reports the performance on this expertly curated benchmark test set. The fine-tuned YOLOv10 model achieves state-of-the-art results, surpassing both the original DocLayout-YOLO and the pre-trained variant across nearly all evaluation metrics. The overall mAP50 reaches 0.981, with mAP50--95 at 0.930, representing a +0.013 gain over the pre-trained model in mAP50. Notably, detection of \textit{Title} blocks improves to 0.986 (mAP50), and \textit{Plain Text} localization achieves near-perfect scores (0.993 mAP50 / 0.978 mAP50--95). Overall, the two-stage training strategy, broad domain adaptation followed by precision fine-tuning on expert-verified data, yields a highly robust layout detector specifically optimized for complex historical newspaper archives.

\begin{table}[h]
    \centering
    \caption{Comparative performance metrics of three YOLO-based models on the benchmark dataset test set.}
    \label{tab:yolo_benchmark_comparison}
    \resizebox{\textwidth}{!}{%
    \begin{tabular}{lcccccccc}
        \toprule
        \textbf{Class} & \textbf{Sample Count} & 
        \multicolumn{2}{c}{\textbf{DocLayout-YOLO}} & 
        \multicolumn{2}{c}{\textbf{Pre-trained YOLOv10}} & 
        \multicolumn{2}{c}{\textbf{Fine-tuned YOLOv10}} \\
        \cmidrule(lr){3-4} \cmidrule(lr){5-6} \cmidrule(lr){7-8}
                       &                       & \textbf{mAP50} & \textbf{mAP50-95} & \textbf{mAP50} & \textbf{mAP50-95} & \textbf{mAP50} & \textbf{mAP50-95} \\
        \midrule
        All            & 641                   & 0.746          & 0.645             & 0.968          & 0.919             & \textbf{0.981} & \textbf{0.930}   \\
        Title          & 99                    & 0.507          & 0.387             & 0.975          & \textbf{0.905}    & \textbf{0.986} & \textbf{0.905}   \\
        Plain Text     & 516                   & 0.912          & 0.819             & 0.987          & 0.968             & \textbf{0.993} & \textbf{0.978}   \\
        Figure         & 12                    & 0.708          & 0.691             & 0.924          & 0.920             & \textbf{0.977} & \textbf{0.958}   \\
        Figure Caption & 14                    & 0.856          & 0.684             & \textbf{0.986} & \textbf{0.884}    & 0.966          & 0.879            \\
        \bottomrule
    \end{tabular}%
    }
\end{table}

\begin{figure}[h!]
    \centering
    
    \includegraphics[width=0.7\linewidth]{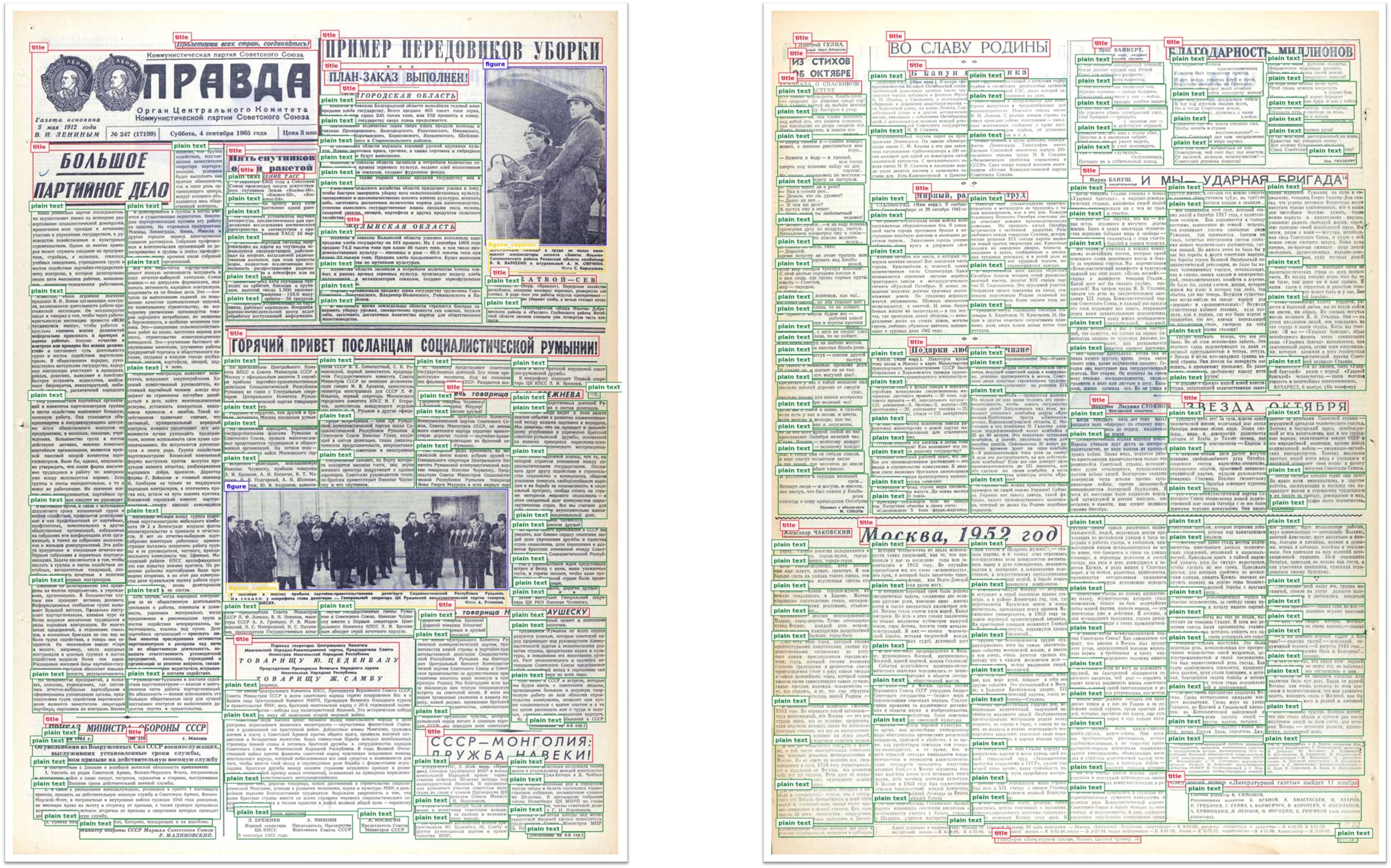} 
    \caption{Fine-tuned YOLOv10 predictions.}
\end{figure}

\section{Text Extraction}

Following layout detection, we extract textual content from each detected region using the Yandex OCR API, which demonstrated superior robustness on our corpus compared to open-source alternatives, particularly in handling the severe visual artifacts characteristic of historical newspapers - including fading, skew, bleed-through, and non-standard fonts prevalent in mid-20th century typography. The API processes high-resolution crops derived from the YOLO-predicted bounding boxes and returns raw text strings with preserved spatial formatting.

To mitigate residual recognition errors without altering the authentic linguistic and typographic character of the source material, we apply a lightweight post-correction stage using YandexGPT 5.1 RC. We adopt a strictly constrained prompting strategy (see Appendix~\ref{app:prompt1}) that directs the model to retain the detected text verbatim—preserving period-appropriate hyphenation while excluding line breaks (\texttt{\textbackslash n}) and correcting only evident OCR artifacts (character substitutions, diacritic errors, and fragmented words). This approach maintains the orthographic and stylistic conventions of the Soviet era (e.g., archaic spellings, punctuation patterns) rather than modernizing the language, ensuring that downstream semantic analysis operates on historically authentic text.

Given that the OCR quality proved sufficiently high for subsequent semantic assembly tasks, with error rates below the threshold required for reliable TF-IDF vectorization and lexical similarity computation, we did not construct a dedicated text-extraction benchmark. Instead, we treat text recognition as a preprocessing component and direct our evaluation efforts toward the core contributions of layout detection and article reconstruction, where annotation ambiguity presents the primary technical challenge.

\section{Semantic Assembly}

Reconstructing semantically coherent articles from disjointed text blocks requires reasoning across both spatial layout and lexical content. While previous sections establish robust detection and recognition of individual blocks, the critical challenge remains determining which fragments belong together and in what order they should be read. Historical newspapers compound this difficulty through irregular column flows, nested headlines, and discontinuous text regions that violate modern linear reading assumptions. We therefore treat semantic reconstruction as a structured inference problem that jointly models visual continuity, geometric relationships, and textual similarity.

\subsection{Problem Formulation} 
Rather than directly predicting reading order, a task confounded by non-linear typographic flows and physical degradation, we reformulate article reconstruction as pairwise binary classification. Given two detected blocks, the classifier determines whether they belong to the same semantic article. From these pairwise affinities, we construct a fully connected graph where edge weights represent linkage confidence. Article-level clusters are recovered via confidence-weighted hierarchical clustering, after which a deterministic left-to-right, top-to-bottom heuristic establishes the final reading sequence. To accommodate layout heterogeneity, we train separate classifiers for distinct association types: (1) \textit{Plain Text $\leftrightarrow$ Plain Text}, and (2) \textit{Title $\leftrightarrow$ Plain Text}, capturing the asymmetric semantic anchoring provided by headlines.

\subsection{Feature Engineering} 
Each candidate block pair is represented by a multi-modal feature vector integrating lexical, spatial, and visual signals:

\begin{itemize}

\item \textbf{Lexical-Semantic Features:} 
OCR-corrected text undergoes morphological lemmatization, lowercasing, and stop-word removal. Given the noise characteristics of historical OCR outputs, we employ character n-gram TF-IDF representations, leveraging their established robustness to recognition artifacts and proven effectiveness for fuzzy lexical matching in degraded text conditions. We utilize character-level n-grams ($n \in [2,4]$) with a 15,000-feature vocabulary, applying document frequency thresholds ($\text{min\_df}=3$, $\text{max\_df}=0.85$) to suppress corpus-specific noise while preserving discriminative semantic patterns.

\item \textbf{Geometric Features:} Twenty spatial descriptors (see Appendix~\ref{app:geom}) encode the geometric relationship between blocks, derived from normalized coordinates $[x_1, y_1, x_2, y_2]$. These include Euclidean and Manhattan distances, relative positional indices (vertical/horizontal precedence), alignment tolerances, aspect ratios, area ratios, and inter-block spacing measurements. For the title-aware classifier, an additional categorical indicator encodes the specific block-type pairing (Title-to-Text vs. Title-to-Title).

\item \textbf{Visual Features:} To capture implicit layout regularities such as column separators, decorative borders, and whitespace continuity, we extract visual context from the region spanning both candidate blocks. The minimal bounding box enclosing the pair is cropped from the page image, resized to YOLOv10 input resolution, and processed through the fine-tuned backbone. Feature maps from the 22nd convolutional layer undergo global average pooling, yielding a 576-dimensional embedding that encodes local visual context unavailable through purely geometric or textual analysis.

\end{itemize}

\subsection{Training Configuration}
We train on the expert-verified benchmark detailed in Section~3.2. For the Plain Text classifier, we construct a substantial balanced training corpus comprising $124\,774$ positive and $124\,774$ negative pairs ($249\,548$ total). Negative pairs are sampled predominantly from distinct articles within the same page to expose the model to realistic layout ambiguities (adjacent columns, parallel stories); cross-page negatives are introduced only upon intra-page exhaustion to prevent structural overfitting while preserving distributional alignment with inference-time conditions.The Title-aware classifier operates over a more constrained relational space, yielding a training set of $30\,210$ positive and $30\,210$ negative pairs ($60\,420$ total), distributed across association types as follows: Title$\leftrightarrow$Plain Text ($56\,126$ pairs), and Title$\leftrightarrow$Title ($4\,294$ pairs). This stratification ensures balanced exposure to asymmetric semantic anchoring patterns.
The Plain Text classifier is evaluated on a held-out test set containing exactly $19\,598$ positive and $19\,598$ negative pairs ($39\,196$ total), while the Title-aware model is assessed on $3\,176$ positive and $3\,176$ negative pairs ($6\,352$ total). This parity eliminates threshold bias in AUC-ROC estimation and enables reliable comparison across model variants.

Both classifiers employ XGBoost with histogram-based tree splitting, configured for high-capacity tabular learning but with task-specific hyperparameters reflecting the differing complexity of their respective association patterns. For each candidate pair, we concatenate the feature vectors of both blocks—comprising lexical-semantic (TF-IDF), geometric, and visual (YOLO embedding) descriptors for each block—into a single high-dimensional representation that captures the joint characteristics of the potential association. The Plain Text classifier uses 300 estimators with a maximum depth of 12 and learning rate 0.03, while the Title-aware classifier employs a deeper ensemble of 400 estimators at depth 10 with a more conservative learning rate of 0.02 and minimum child weight of 2 to regularize the higher-dimensional feature space. Both models apply subsample and column sampling ratios of 0.8, with optimization targeting AUC-ROC under binary logistic loss.

\subsection{Results}
\begin{figure}[H]
    \centering
    \includegraphics[width=0.65\linewidth]{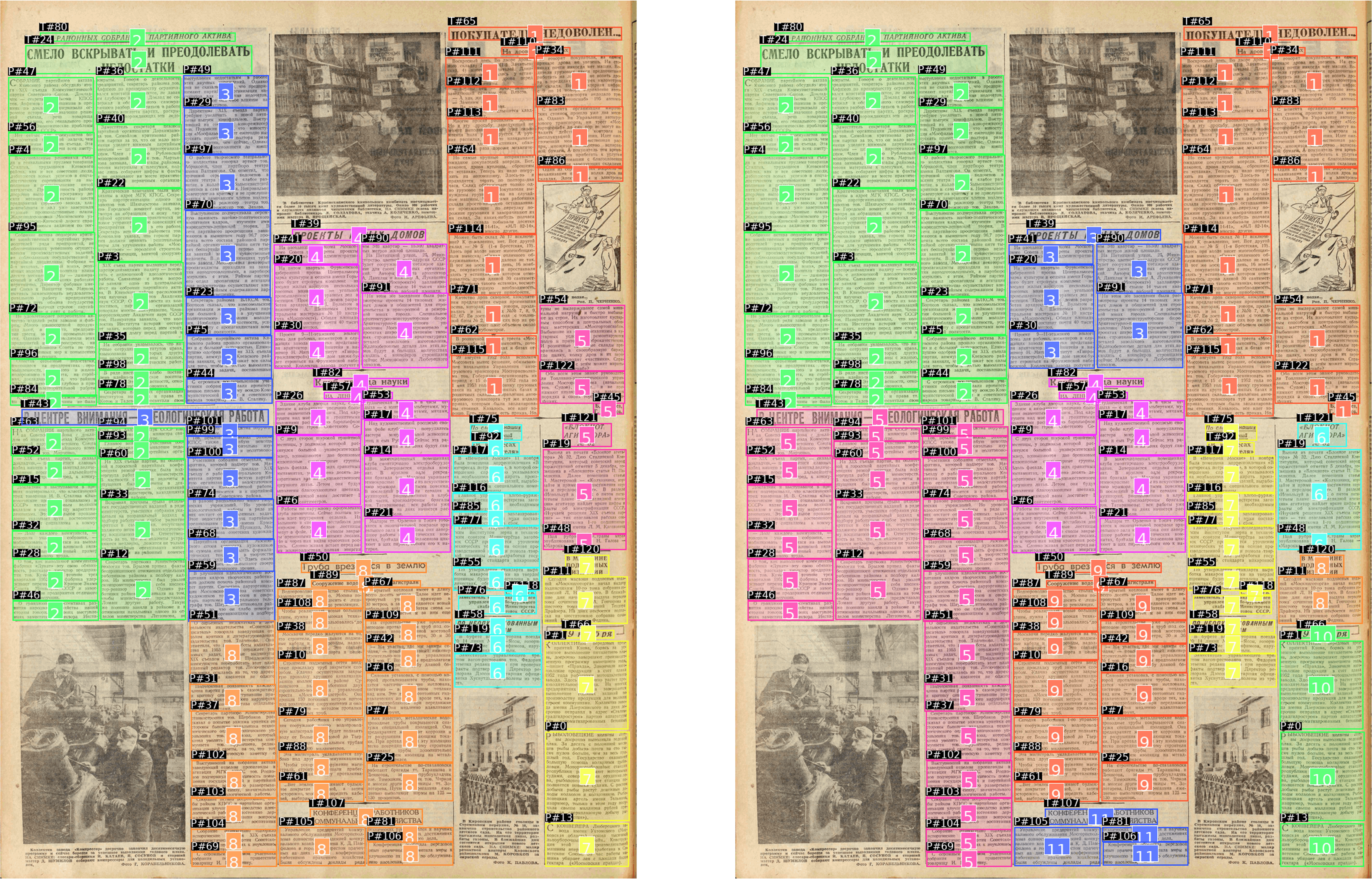} 
    \caption{Comparison of article reconstruction results. \textbf{Left:} Baseline predictions without multi-modal YOLO embeddings. \textbf{Right:} Final \textbf{PereStruct} predictions integrating visual embeddings. Color mapping corresponds to distinct semantic articles.}
    \label{fig:final_predictions}
\end{figure}

A baseline architecture employing only TF-IDF lexical features and geometric coordinate derivatives achieves moderate discriminative performance (Table~\ref{tab:baseline_semantic}). The Plain Text classifier yields F1$=$0.743, while the Title-aware model reaches F1$=$0.882. This performance gap reflects the inherent asymmetry of the association tasks: title-to-text linkages are strongly constrained by vertical proximity and typographic hierarchy, making them readily predictable even without visual context. Conversely, plain text-to-text associations present substantially greater ambiguity, as disjointed blocks from the same article may be separated by arbitrary spatial gaps, intervening columns, or complex multi-flow layouts, causing the baseline to frequently fail on scattered fragments (Figure~\ref{fig:final_predictions}, left).

\begin{table}[h]
    \centering
    \caption{Baseline performance using TF-IDF and geometric features only.}
    \label{tab:baseline_semantic}
    \begin{tabular}{lccc}
        \toprule
        \textbf{Classifier} & \textbf{AUC} & \textbf{Acc} & \textbf{F1} \\
        \midrule
        Plain Text $\leftrightarrow$ Plain Text & 0.851 & 0.774 & 0.743 \\
        Title $\leftrightarrow$ \{Plain Text, Title\} & 0.941 & 0.885 & 0.882 \\
        \bottomrule
    \end{tabular}
\end{table}

To capture implicit spatial relationships and visual continuity cues -- such as column separators, whitespace patterns, and decorative borders between blocks -- we augment the feature space with fine-grained visual embeddings. For each candidate pair, we extract a tight bounding box encompassing both blocks (with 5-pixel padding to include interstitial context), resize the crop to $1024$ px to match the YOLOv10 input resolution, and forward it through the fine-tuned backbone. We extract feature maps from the 22nd convolutional layer and apply global average pooling across spatial dimensions, yielding a compact 576-dimensional dense embedding that encodes local visual geometry unavailable through coordinate features alone.

Incorporating these YOLO-derived visual embeddings substantially bridges the performance gap (Table~\ref{tab:multimodal_results}). The Plain Text classifier improves to \textbf{F1$=$0.850}, demonstrating that visual continuity signals effectively resolve ambiguous spatial adjacencies and implicit section boundaries. The Title-aware model achieves \textbf{F1$=$0.904}, approaching optimal performance (Figure~\ref{fig:final_predictions}, right). These high-confidence pairwise affinities enable robust graph-based clustering, yielding complete semantic article reconstructions that generalize across complex historical layouts with minimal heuristic post-processing.

\begin{table}[h]
    \centering
    \caption{Multi-modal performance with YOLO visual embeddings.}
    \label{tab:multimodal_results}
    \begin{tabular}{lccc}
        \toprule
        \textbf{Classifier} & \textbf{AUC} & \textbf{Acc} & \textbf{F1} \\
        \midrule
        Plain Text $\leftrightarrow$ Plain Text & \textbf{0.925} & \textbf{0.859} & \textbf{0.850} \\
        Title $\leftrightarrow$ \{Plain Text, Title\} & \textbf{0.961} & \textbf{0.907} & \textbf{0.904} \\
        \bottomrule
    \end{tabular}
\end{table}

\section{Evaluation}
\label{sec:evaluation}

To validate the effectiveness of our proposed approach, we evaluate PereStruct on end-to-end document understanding against state-of-the-art vision-language models (VLMs). We compare our modular pipeline against two strong baselines: Qwen3.6-35B-A3B (a 35B parameter model) and Qwen3.6-Plus (a larger-scale proprietary model), representing current advances in multimodal large language models.

\subsection{Experimental Setup}

\textbf{Task Definition.} Models are evaluated on the task of converting raw newspaper page images into structured Markdown format containing all articles grouped in human reading order (left-to-right, top-to-bottom). This requires both accurate text recognition and correct logical document structure reconstruction.

\textbf{Dataset.} We evaluate on the held-out test set of PereStruct-Benchmark ($N=10$ pages), comprising complex Soviet-era newspaper layouts with dense text, multi-column formatting, and degraded image quality due to age. These pages were strictly excluded from training both the YOLO detector and the XGBoost classifier.

\textbf{Baselines.} For Qwen models, we utilize the same prompt across all variants (see Appendix~\ref{app:prompt}), requesting Markdown output with hierarchical headings and preserved formatting. All Qwen experiments use ``fast'' thinking mode (thinking budget disabled). To ensure fair comparison, generated outputs are standardized by stripping Markdown formatting tokens and newline characters prior to metric calculation.

\subsection{Evaluation Metrics}

We employ four complementary metrics to assess both local text accuracy and global structural fidelity:

\begin{itemize}
    \item \textbf{BLEU}~\cite{papineni2002bleu}: Quantifies n-gram precision between generated and reference texts, capturing the correctness of text extraction and preservation of local formatting structures.
    \item \textbf{ROUGE-1}~\cite{lin2004rouge}: Measures unigram overlap, assessing word-level content coverage and terminology accuracy.
    \item \textbf{ROUGE-2}~\cite{lin2004rouge}: Evaluates bigram overlap, sensitive to local phrase construction and adjacent word ordering.
    \item \textbf{ROUGE-L}~\cite{lin2004rouge}: Computes the longest common subsequence, evaluating global document structure and the correct reconstruction of logical reading order.
\end{itemize}

\subsection{Results and Analysis}

Table~\ref{tab:main_results} presents the quantitative comparison. PereStruct substantially outperforms both VLM baselines across all metrics, achieving mean scores of $0.959$ (BLEU), $0.961$ (ROUGE-1), $0.912$ (ROUGE-2), and $0.916$ (ROUGE-L).

\begin{table}[htbp]
\centering
\caption{End-to-end performance comparison on PereStruct-Benchmark test set. Higher is better for all metrics.}
\label{tab:main_results}
\begin{tabular}{@{}lcccc@{}}
\toprule
\textbf{Model} & \textbf{BLEU} $\uparrow$ & \textbf{ROUGE-1} $\uparrow$ & \textbf{ROUGE-2} $\uparrow$ & \textbf{ROUGE-L} $\uparrow$ \\
\midrule
Qwen3.6-35B-A3B & 0.119 & 0.394 & 0.174 & 0.284 \\
Qwen3.6-Plus & 0.337 & 0.628 & 0.366 & 0.488 \\
\midrule
\textbf{PereStruct} & \textbf{0.959} & \textbf{0.961} & \textbf{0.912} & \textbf{0.916} \\
\bottomrule
\end{tabular}
\end{table}

The 35B parameter Qwen model struggles significantly with BLEU scores below $0.12$, indicating severe token-level hallucinations. Despite its larger scale and 1M token context window, Qwen3.6-Plus achieves only moderate performance (ROUGE-L: $0.488$), failing to reliably reconstruct document structure.

\textbf{Error Analysis.} We observe that VLMs exhibit systematic failure modes when processing complex historical layouts:

(1) \textit{Structural hallucination} - generating plausible but non-existent text when character recognition fails;

(2) \textit{Repetition artifacts} - repeating phrases instead of parsing distinct blocks;

(3) \textit{Layout confusion} - merging adjacent articles or splitting single articles due to dense columnar formatting. The high ROUGE-L gap ($0.916$ vs $0.488$) specifically highlights PereStruct's superior ability to maintain logical reading order through explicit block detection and pairwise classification.

These results demonstrate that end-to-end VLM approaches, despite advances in scale and context length, remain insufficient for high-stakes document digitization tasks requiring perfect structural fidelity. PereStruct's modular architecture combining specialized components for detection (YOLO), recognition (OCR), and semantic assembly (XGBoost) provides the necessary reliability for archival document processing.
 
\section{Conclusion}

We present PereStruct, a modular pipeline for high-fidelity parsing of historical newspapers that addresses the persistent challenge of semantic reconstruction from degraded, complex archival layouts. By integrating fine-tuned layout detection with a novel multi-modal assembly module, our approach bridges the domain gap that limits generic vision-language models on out-of-distribution historical documents.

The key technical advance lies in our semantic assembly framework, which jointly leverages lexical-semantic similarity, geometric constraints, and fine-grained visual embeddings extracted from the layout detection backbone. This multi-modal fusion proves critical for resolving ambiguous associations in irregular multi-column layouts, where purely textual or spatial cues prove insufficient. Our empirical results demonstrate that incorporating YOLO-derived visual features substantially improves pairwise classification performance, lifting the \textit{Plain Text} classifier F1 from 0.743 to \textbf{0.850} and achieving \textbf{F1=0.904} for title-aware associations—enabling robust graph-based reconstruction of semantically coherent articles with minimal post-processing heuristics.

Crucially, our end-to-end evaluation reveals that even state-of-the-art vision-language models struggle with this task: despite their billion-scale parameters and million-token contexts, both Qwen3.6-35B-A3B and Qwen3.6-Plus exhibit severe hallucinations and structural errors on complex historical layouts, achieving BLEU scores below 0.34 compared to PereStruct's 0.96. These findings validate that modular, task-specific architectures remain essential for high-stakes document digitization, where generic end-to-end approaches fail to reliably preserve both textual accuracy and logical reading order.

Beyond algorithmic contributions, we release the PereStruct resources to the research community: a corpus of 599 human-annotated pages for layout detection and a rigorously curated \textbf{PereStruct} benchmark of 93 pages with expert-verified block-to-article mappings and OCR-corrected ground truth. These datasets provide the first standardized evaluation framework specifically designed for end-to-end historical newspaper parsing, lowering barriers for future research in archival document understanding.

While our current implementation focuses on Soviet-era newspapers, the modular architecture—separating layout detection, text extraction, and semantic assembly—generalizes to diverse historical corpora. We anticipate that PereStruct will serve as a foundation for large-scale digitization efforts, enabling historians and archivists to unlock structured access to vast collections of visually complex historical documents.

\section*{Acknowledgments}
The article was prepared with the support of the Center for Technologies for Society of the Yandex Cloud platform. The center provided access to computing resources based on a grant from the Science and Education Support Program.

We thank the Yandex Crowd team for their hard work!

We thank the OpenDataLab team for the great work on DocLayout-YOLO!

\bibliographystyle{unsrt}  
\bibliography{references}  

\appendix

\section{OCR Post-Correction Prompt}
\label{app:prompt1}

The following constrained prompt was used with YandexGPT 5.1 RC to mitigate OCR artifacts while preserving period-specific linguistic characteristics:

\begin{otherlanguage}{russian}
\begin{verbatim}
Ты — эксперт по русскому языку и историческим текстам, особенно по материалам советской эпохи. 
Твоя задача — исправить только явные опечатки, возникшие в результате ошибок OCR 
(распознавания текста с изображений), не внося никаких других изменений в текст.

Правила:

Не перефразируй, не улучшай стиль, не «современизируй» лексику и не 
исправляй грамматические конструкции, характерные для советской эпохи.

Сохраняй оригинальную пунктуацию, регистр букв, орфографию и стилистику, 
даже если они кажутся устаревшими или не соответствуют современным нормам.

Исправляй только те символы или слова, которые очевидно искажены OCR 
(например: «сегоднл» → «сегодня», «ро6очий» → «рабочий», «пoддержкa» → «поддержка»).

Если сомневаешься — оставляй как есть.

Не добавляй, не удаляй и не переставляй слова, предложения или абзацы.

Ты должен удалить \n и исправить переносы слов (пример: при-\n своенный -> присвоенный).

Никогда не меняй регистр букв (с маленькой на заглавную и наоборот).

Верни только исправленный текст, без пояснений, комментариев или форматирования.
\end{verbatim}
\end{otherlanguage}

\section{Geometric Features}
\label{app:geom}

The geometric feature vector encodes pairwise spatial relationships between candidate blocks. For two blocks $b_1$ and $b_2$ with normalized coordinates $c_1 = (x_1, y_1, x_{1b}, y_{1b})$ and $c_2 = (x_2, y_2, x_{2b}, y_{2b})$, we compute $w_i = x_{ib} - x_i$, $h_i = y_{ib} - y_i$ (widths and heights), and center coordinates $(cx_i, cy_i)$. The resulting 20-dimensional feature vector comprises:

\begin{enumerate}[leftmargin=*]
    \item \textbf{Euclidean distance:} $\sqrt{(cx_1 - cx_2)^2 + (cy_1 - cy_2)^2}$
    \item \textbf{Manhattan distance:} $|cx_1 - cx_2| + |cy_1 - cy_2|$
    \item \textbf{Chebyshev distance:} $\max(|cx_1 - cx_2|, |cy_1 - cy_2|)$
    \item \textbf{Absolute horizontal offset:} $|cx_1 - cx_2|$
    \item \textbf{Absolute vertical offset:} $|cy_1 - cy_2|$
    \item \textbf{Vertical ordering indicator:} $\mathbb{I}(cy_1 < cy_2)$ (1 if $b_1$ is above $b_2$)
    \item \textbf{Horizontal ordering indicator:} $\mathbb{I}(cx_1 < cx_2)$ (1 if $b_1$ is left of $b_2$)
    \item \textbf{Quadrant encoding:} $2 \cdot \mathbb{I}(cx_1 > cx_2) + \mathbb{I}(cy_1 > cy_2)$ (encodes relative position in 4 quadrants)
    \item \textbf{Horizontal alignment:} $\mathbb{I}(|cx_1 - cx_2| < \min(w_1, w_2)/2)$ (1 if centers align horizontally within half the narrower width)
    \item \textbf{Vertical alignment:} $\mathbb{I}(|cy_1 - cy_2| < \min(h_1, h_2)/2)$ (1 if centers align vertically within half the shorter height)
    \item \textbf{Width of block 1:} $w_1$
    \item \textbf{Width of block 2:} $w_2$
    \item \textbf{Height of block 1:} $h_1$
    \item \textbf{Height of block 2:} $h_2$
    \item \textbf{Width ratio:} $w_1 / (w_2 + \epsilon)$ (aspect ratio scaling)
    \item \textbf{Height ratio:} $h_1 / (h_2 + \epsilon)$
    \item \textbf{Absolute area difference:} $|w_1 h_1 - w_2 h_2|$
    \item \textbf{Area ratio:} $(w_1 h_1) / (w_2 h_2 + \epsilon)$
    \item \textbf{Horizontal gap:} $\max(0, x_2 - x_{1b})$ if disjoint to the right, else $\max(0, x_1 - x_{2b})$ (0 if overlapping)
    \item \textbf{Vertical gap:} $\max(0, y_2 - y_{1b})$ if disjoint below, else $\max(0, y_1 - y_{2b})$ (0 if overlapping)
\end{enumerate}

where $\epsilon = 10^{-8}$ ensures numerical stability. These features capture relative positioning (1--5), reading-order heuristics (6--8), alignment conditions (9--10), dimensional similarity (11--18), and spatial separation (19--20) necessary for determining logical article adjacency in complex multi-column layouts.

\section{Qwen Vision-Language Model Prompt}
\label{app:prompt}

The following Russian-language prompt was used for all Qwen model evaluations, requesting structured Markdown extraction while preserving historical stylistic elements:

\begin{otherlanguage}{russian}
\begin{verbatim}
Ты — эксперт по архивным документам эпохи СССР. Извлеки все статьи из приведенного изображения
газетной страницы.

Правила извлечения:
1. Читай страницу по колонкам: сначала левую колонку сверху вниз, затем правую.
2. Для каждой статьи создай блок в формате Markdown.
3. Первый заголовок статьи оформляй как "# Заголовок".
4. Все последующие подзаголовки внутри статьи оформляй как "## Подзаголовок".
5. Основной текст сохраняй с точным переносом строк (\n), как на изображении.
6. Игнорируй: таблицы, колонтитулы (шапку/низ страницы), подписи к рисункам, рекламу.
7. Игнорируй заголовки, которые не относятся к статьям (название газеты, даты, № газеты).
8. Сохраняй стилистику советской эпохи: слова типа "товарищ", "бригада", "пятилетка", "комсомол".
9. Сохраняй орфографию и пунктуацию оригинала (включая Е/Ё, дефисы в словах).

Выводи только статьи в чистом виде без лишних комментариев.
\end{verbatim}
\end{otherlanguage}

\end{document}